\documentclass[conference]{IEEEtran}
\IEEEoverridecommandlockouts
\usepackage{cite}
\usepackage{pifont}       
\usepackage{bbding}       
\usepackage{fontawesome}  

\usepackage{amsmath,amssymb,amsfonts}
\usepackage{algorithmic}
\usepackage{graphicx}
\usepackage{textcomp}
\usepackage{color}

\usepackage[normalem]{ulem}
\usepackage{multirow}
\useunder{\uline}{\ul}{}
\def\BibTeX{{\rm B\kern-.05em{\sc i\kern-.025em b}\kern-.08em
    T\kern-.1667em\lower.7ex\hbox{E}\kern-.125emX}}

\newcommand{\cmark}{\ding{51}}
\newcommand{\xmark}{\ding{55}}

\begin{document}

\title{

TLAG: An Informative \textbf{T}rigger and \textbf{L}abel-\textbf{A}ware Knowledge \textbf{G}uided Model for Dialogue-based Relation Extraction

\thanks{* Corresponding author.}
}

\author{
    \IEEEauthorblockN{Hao An$^{}$, Dongsheng Chen$^{}$, Weiyuan Xu$^{}$, Zhihong Zhu$^{}$, Yuexian Zou$^{*}$}
    \IEEEauthorblockA{$^{}$ ADSPLAB, School of ECE, Peking University, Shenzhen, China}
    \IEEEauthorblockA{\{anhao, chends, xuwy, zhihongzhu\}@stu.pku.edu.cn, zouyx@pku.edu.cn}
}

\maketitle

\begin{abstract}
Dialogue-based Relation Extraction (DRE) aims to predict the relation type of argument pairs that are mentioned in dialogue. The latest trigger-enhanced methods propose trigger prediction tasks to promote DRE. However, these methods are not able to fully leverage the trigger information and even bring noise to relation extraction. To solve these problems, we propose TLAG, which fully leverages the trigger and label-aware knowledge to guide the relation extraction. First, we design an adaptive trigger fusion module to fully leverage the trigger information. Then, we introduce label-aware knowledge to further promote our model's performance. Experimental results on the DialogRE dataset show that our TLAG outperforms the baseline models, and detailed analyses demonstrate the effectiveness of our approach.

\end{abstract}


\section{Introduction}
Relation extraction (RE) aims to identify the semantic relation between two entities mentioned in unstructured text. The results of RE contain relational information that benefits a variety of natural language processing (NLP) applications, such as knowledge graph construction~\cite{KG_survey}, question answering~\cite{RE4QA}, dialogue system~\cite{RE_DS_survey}, digital twins~
\cite{RE_DS_survey} and so on. Previous studies mainly focus on extracting relations within a single sentence and achieve remarkable success. Despite their success, sentence-level RE suffers from an inevitable restriction in practice: a large number of relational facts are expressed in complex contexts composed of multiple sentences~\cite{DocRED, Complex_relation_extraction_Challenges_opportunities, More_Data}. Therefore, extracting relations across sentences is necessary.

\begin{figure}[!htbp]
\centerline{\includegraphics[width=0.95\linewidth]{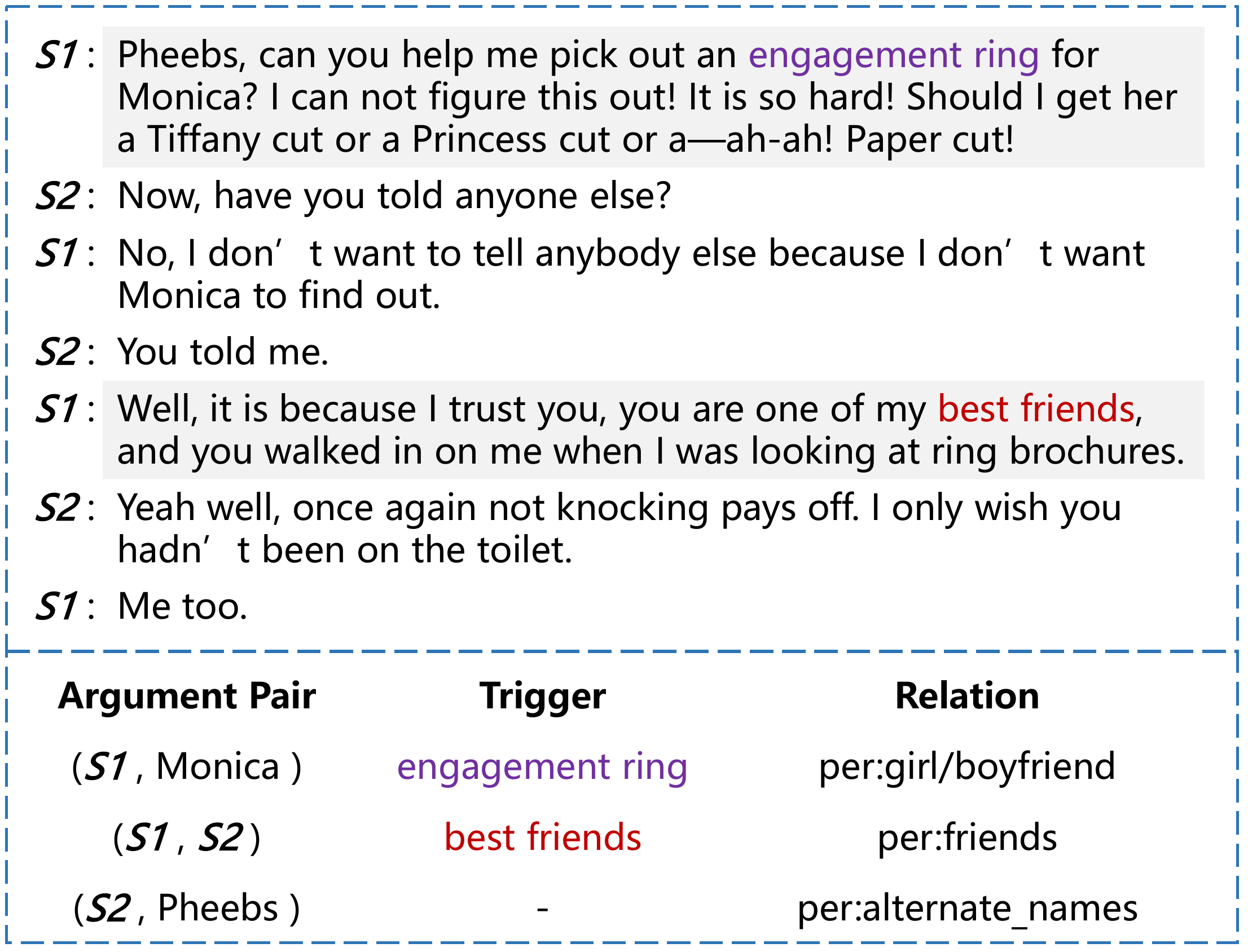}}
\caption{An example of dialogue-based relation extraction; S1 and S2 are anonymized speaker of each utterance. Triggers are clues of relations; Relations and corresponding triggers of two pairs of entities are annotated.}
\label{data_example}
\end{figure}

Recently, dialogue-based relation extraction (DRE) has raised much research interest, which aims to classify the relation type between an argument pair within a multi-turn dialogue. As shown in Figure~\ref{data_example}, given a dialogue and an argument pair, DRE aims to identify the relationship between the two arguments, and the trigger is the most informative word or phrase that indicates the relation type. In this case, “\textit{engagement ring}” is the trigger span between \textit{S1} and “\textit{Monica}”, which indicates a “\textit{per: girl/boyfriend}” relation.

Compared with a single utterance, a dialogue paragraph has the characteristics of low information density~\cite{Low-information-density}, which naturally brings challenges to the research of DRE. The dialogue example shown in Figure~\ref{data_example} includes seven turns of utterances, however, only two utterances among them are helpful to the final relation extraction. Existing methods, including TWP~\cite{Trigger_Words_Prediction} and TREND~\cite{TREND}, leverage the trigger information via trigger span prediction to solve this problem. However, these approaches still have two shortcomings. First, they do not fully leverage the trigger information. TWP does not feed the trigger features to the relation classification module. TREND fuses the trigger with the dialogue context feature via the vanilla attention mechanism, which ignores the mutual information between the argument pair. Second, TREND adopts a simple extractive module as the trigger predictor. However, the predictor does not perform well in the dialogue dataset, which brings noise to relation classification when the predicted trigger is wrong.

To this end, we propose a novel approach with adaptive trigger fusion and label-aware knowledge guidance, which takes full advantage of trigger and the knowledge. Our main contributions are threefold:
\begin{itemize}
    \item We propose a novel adaptive trigger fusion mechanism to fuse the trigger information with context information and argument information. Thus a context-aware and argument-aware trigger feature can be obtained.
    \item We leverage label-aware semantic knowledge to guide the trigger extraction. Therefore, we promote the performance of the trigger extractor and reduce the noise effect on relation classification.
    \item We conduct a series of comparative experiments on public DRE dataset. Experimental results show that our TLAG outperforms DRE baseline models. The detailed analyses show that our approach is able to leverage the trigger and the label-aware knowledge efficiently.
\end{itemize}

The rest of this paper is organized as follows: Section~\ref{related_work} summarizes the related work; Section~\ref{Section_Method} shows the overview and details of our approach; Section~\ref{Section_experiments} provides the experimental results and detailed discussion; Section~\ref{section_conclusion} gives a conclusion for our work.

\section{Related Works}\label{related_work}
Relation extraction (RE) has been studied widely over the past decade and many approaches have achieved remarkable success. Most of the earlier approaches focus on sentence-level relation extraction, including CNN-based method~\cite{CNNRE}, dependency-based method~\cite{DependencyRE}, LSTM-based~\cite{LSTMRE} method, and attention-based~\cite{AttentionRE} method. However, most relational facts can only be extracted from complicated contexts like dialogues rather than single sentences~\cite{DocRED}, which should not be neglected.

Recently, dialogue-based relation extraction (DRE), which aims to extract relations in multi-turn dialogues, has attracted increasing research interest. The DRE models are mainly divided into two categories: sequence-based model and graph-based model.

The sequence-based models encode the dialogue as a long sequence using LSTM~\cite{LSTM} or BERT~\cite{BERT} and then employ a series of complex mechanisms, such as attention mechanisms and gate mechanisms, to capture key information in the dialogue. Yu et al.~\cite{DRE} adapt the BERT model to the dialogue setting by inserting special tokens. Xue et al.~\cite{SimpleRE} propose a simple and efficient relation refinement mechanism that reports good performance. However, these models handle the multi-turn dialogue by concatenating all the utterances, not taking the dialogue-specific characteristics into account.

The graph-based models build a graph by connecting the nodes in different ways. In general, each node in graph represents a token, an utterance, or an entity in the given dialogue. Chen et al.~\cite{DHGAT} build a heterogeneous graph attention network that models multi-type features of the conversation, such as words, speakers, entities, entity types, and utterances nodes. Nan et al.~\cite{LSR} construct meta dependency paths of each argument pair and aggregate all the word representations located in these paths to their model to enhance the model’s reasoning ability. Xue et al.~\cite{GDPNet} propose to construct a latent multi-view graph to capture various possible relationships among tokens. Lee and Choi~\cite{TUCORE} introduce a heterogeneous dialogue graph to model the interaction among nodes (e.g., speakers, utterances, arguments) and propose a GCN mechanism combined with contextualized representations of turns.

Recent studies leverage the trigger information to promote dialogue-based relation extraction. Zhao et al.~\cite{Trigger_Words_Prediction} 
propose two auxiliary tasks, a speaker prediction task that captures the characteristics of speaker-related entities, and a trigger words prediction task that provides supportive contexts for relations. Lin et al.~\cite{TREND} propose an end-to-end model with an explicit trigger gate for trigger existence, an extractive trigger predictor, and an attention feature fusion, which leverages the promotes DRE via fusing trigger information. However, these approaches are not able to fully leverage the trigger information.
\section{Method}\label{Section_Method}
\begin{figure*}[htbp!]
\centerline{\includegraphics[width=0.8\linewidth]{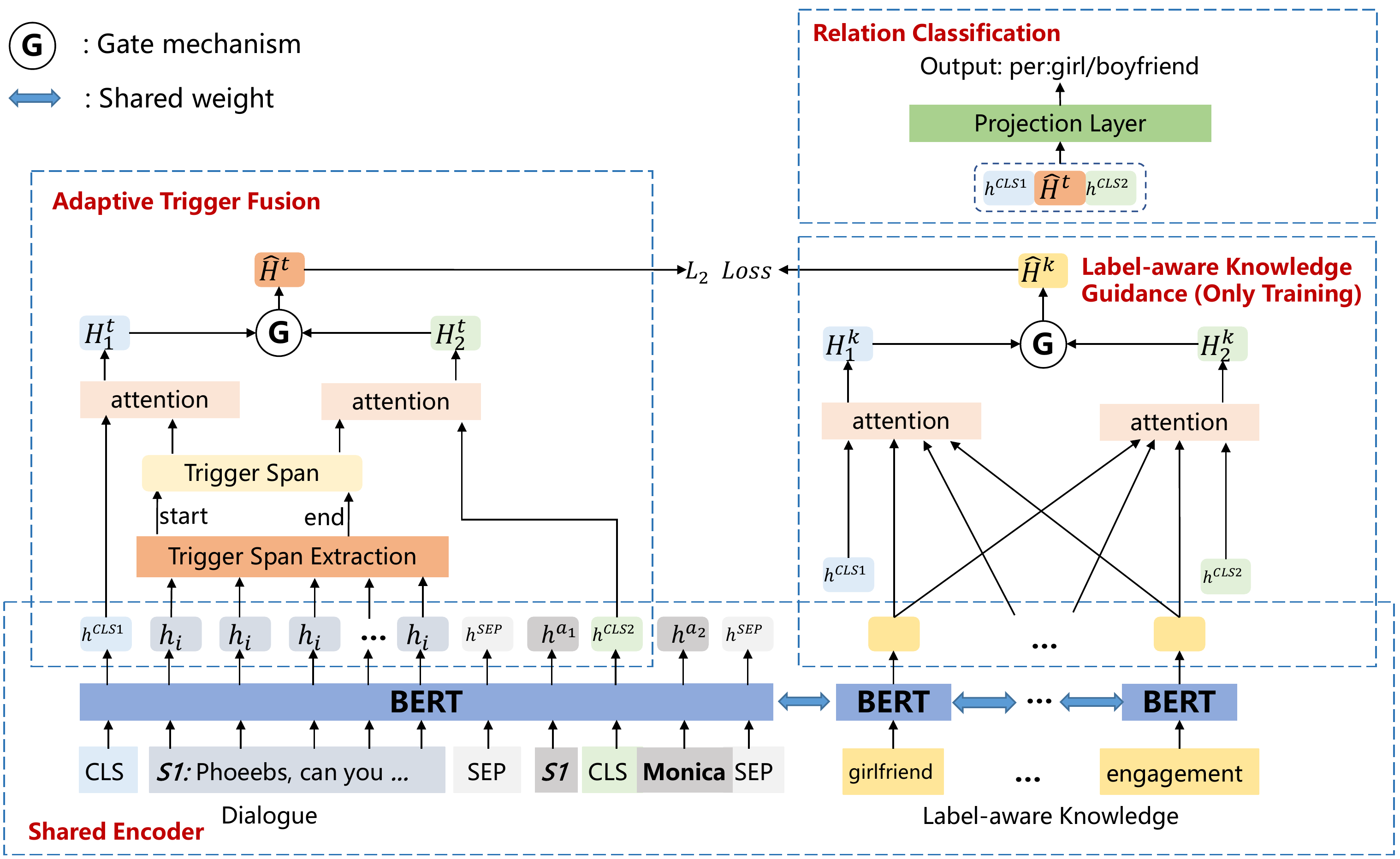}}
\caption{The overall architecture of proposed TLAG. It consists of four modules, i.e., shared encoder, adaptive trigger fusion, label-aware knowledge guidance, and relation classification. The label-aware knowledge guidance is only used for training. }
\label{model}
\end{figure*}

In this section, we first formulate the dialogue-based relation extraction task, then describe the proposed DRE model shown in Figure~\ref{model}. 


\subsection{Task Formulation}
Given a dialogue context $\mathcal{D}$ and a query argument pair composed of a subject entity and an object entity $a=(a_1,a_2)$, the goal is to identify the relation type $r$ between the given argument pair from a predefined relation set $\mathcal{R}$.

\subsection{Shared Encoder}
Following the sequence-based works, we utilize the pre-trained language model BERT~\cite{BERT} as the encoder to yield the contextualized representations for the dialogue and argument pair. And we follow the preprocessing method of BERTs~\cite{DRE} to process the dialogue text. Given a dialogue $\mathcal{D}=s_1:u_1,s_2:u_2,...,s_n:u_n$ with an argument pair $a=(a_1,\; a_2)$, where $s_i$ and $u_i$ denote the speaker ID and the $i^{th}$ turn of the dialogue, and $n$ is the total number of turns, we construct $\hat{\mathcal{D}}=\hat{s}_1:u_1,\hat{s}_2:u_2,...,\hat{s}_n:u_n$, where $\hat{s}_i$ is:
\begin{equation}
    \hat{s}_i =
\begin{cases}
\lbrack{S_1}\rbrack,& {\text{if}\; s_i = a_1}\\
\lbrack{S_2}\rbrack,& {\text{if}\; s_i = a_2}\\
\;\;s_i,& {\text{otherwise}}
\end{cases}
\end{equation}
where $\lbrack{S_1}\rbrack$ and $\lbrack{S_2}\rbrack$ are two newly-introduced special tokens. Besides, we define $\hat{a}_k(k \in \{1,2\})$ to be $[S_k]$ if $\exists{i}(s_i\,=\,a_k)$, and $a_k$ otherwise. Then we append the arguments pair the the sequence: 
\begin{equation}
    input = [[CLS], \hat{\mathcal{D}} , [SEP], \hat{a}_1, [CLS], \hat{a}_2],
\end{equation}
where the first $[CLS]$ token encodes the global dialogue context, and the second one is used to learn the mutual information between argument pair $(a_1, a_2)$, the $[SEP]$ is a separator token. We feed the sequence into BERT and obtain the hidden contextualized representation of each input token. Among them, $h^{CLS1} \in{\mathbb{R}^{d_h}}$ and $h^{CLS2} \in{\mathbb{R}^{d_h}}$ are the hidden output of the two $[CLS]$ tokens respectively, where $d_h$ is the hidden size of BERT.
\subsection{Adaptive Trigger Fusion}\label{gate}
To identify the relation, it is necessary to utilize the trigger information in the dialogue text. Yu et al.~\cite{DRE} have verified that triggers play an essential role in DRE. Inspired by previous methods that use dynamic gate mechanism~\cite{HOA}\cite{LiteVL}, we propose a novel adaptive information fusion mechanism. It not only leverages the trigger information to guide the relation classification but also serves as an effective way to inject knowledge.
\subsubsection{Trigger Span Extraction}\label{trigger_span_extract}
Since the trigger words are only provided during training in practice, we first adopt a trigger span extraction task~\cite{TREND} to predict the trigger span, which is an extractive method prevalent in extractive question answering~\cite{SpanPrediction}. Specially, we perform start-end pointer prediction over the last hidden states of BERT:
\begin{equation}
    \begin{split}
        h^{S} &= W^{start}h + b^{start},\\
        h^{E} &= W^{end}h + b^{end},
    \end{split}
\end{equation}
where $W^{start}, W^{end} \in \mathbb{R}^{d_h}$; $b^{start}, b^{end} \in \mathbb{R}$ are trainable parameters. The cross-entropy loss is used to train the trigger extractor:
\begin{equation}
    \begin{split}
    p(start=i, end=j) = softmax(h^{S}_{i} h^{E}_{j}),\\
         \mathcal{L}_{t}= CELoss({start}, start^{\prime}) + CELoss({end}, end^{\prime}),
    \end{split}
\end{equation}
where ${start'}$ and ${end'}$ are the predicted start and end pointer of the trigger span, and ${start}$ and ${end}$ are the corresponding label.
\subsubsection{Adaptive Gate Fusion}

After obtaining the predicted trigger span $t=[t_1,t_2,...,t_m]$, we introduce a novel adaptive fusion module to obtain the context-sensitive and entity-sensitive features representation of the trigger. The fusion process includes the following two steps. 

Firstly, we apply a parameter-free attention mechanism to obtain context-aware and argument-aware trigger feature:
\begin{equation}  
    \begin{split}
        \alpha_{i}^1&= softmax(t_i \cdot h^{CLS1}),\\
        H_1^t&= \alpha_{i}^{1} \cdot h^{CLS1},\\
        \alpha_{i}^2&=softmax(t_i \cdot h^{CLS2}),\\
        H_{2}^t&= \alpha_{i}^2 \cdot h^{CLS2}.
    \end{split}
\end{equation}
Secondly, to adaptively fuse useful information from context-aware trigger feature $H_1^t$ and argument-aware trigger feature $H_2^t$, we introduce our gate fusion module:
\begin{equation}  
    \begin{split}
        G_t &= \sigma(\mu_t),   \\
        \hat{H}^{t}&=G_t H_1^t + (1-G_t)H_2^t,
    \end{split}
\end{equation}
where $\mu_t \in {\mathbb{R}^{d_h}}$ is the introduced learnable parameter, $\sigma$ is the sigmoid activation function and $G_t$ is the gate that determines the weight of these two distinct features. $\hat{H}^{t}$ is the final fused semantic feature and it will be propagated to the knowledge guidance module and the relation classification head.

\subsection{Label-aware Knowledge Guidance}
In the adaptive trigger fusion module, we obtain the trigger span $t=[t_1,t_2,...,t_m]$ by the trigger extraction task. However, an incorrect trigger span brings noise to the model and results in performance degradation, thus it is crucial to obtain the correct trigger. Inspired by the knowledge-enhanced pre-trained language model~\cite{ERNIE}, and the success of leveraging the label information to optimize the model~\cite{zhu}, we utilize label semantic knowledge to guide the trigger extractor.

For each relation $r$ in $\mathcal{R}$, we obtain five most relevant words from concept graph~\cite{concept_graph}, as well as the top-5 trigger words in the DialogRE train set. For example, the knowledge set of the relation \emph{“per: girl/boyfriend”} is \emph{\{girlfriend, boyfriend, engagement, marry, relationship, wedding, lover, love, couple, together\}}. We denotes the knowledge set as $K^r=\{K^r_1, K^r_2,...,K^r_{r_n}\}$, then we feed the knowledge into a siamese BERT encoder~\cite{Siamese_TEXT_MATCH} to get hidden representation of knowledge:
\begin{equation}  
    \begin{split}
        h^{K_i^r}&=BERT(K^r_i)\;(1\le i \le r_n).
    \end{split}
\end{equation}
Since the knowledge and the dialogue are not encoded in the same context, we can not utilize the knowledge feature directly. So the semantic transformation is performed by the adaptive trigger fusion mechanism proposed in subsection~\ref{gate}. 

We first fuse the knowledge with the context feature $h^{CLS1}$ and argument feature $h^{CLS2}$ respectively:
\begin{equation}
    \begin{split}
        H_{1}^k&= max(h^{K_i^r} \cdot h^{CLS1}),\ 1\le i \le r_n\\
        H_{2}^k&= max(h^{K_i^r} \cdot h^{CLS2}). \, \ 1\le i \le r_n
    \end{split}
\end{equation}
where $H_{1}^k$ is context-aware feature and $H_{2}^k$ is argument-aware feature. Then we adopt the adaptive gate to fuse the context-aware and argument-aware features:
\begin{equation}
    \begin{split}
        G_k &= \sigma(\mu_k),\\
        \hat{H}^k&=G_k H^{k}_{1} + (1-G_t)H^k_2,
    \end{split}
\end{equation}
where $\mu_k \in {\mathbb{R}^{d_h}}$ is a learnable parameter, $\sigma$ is the sigmoid activation function and $\hat{H}^k$ is the fused label knowledge feature. Finally, we leverage $\hat{H}^k$ to guide the trigger span extraction by $L_2$ loss:
\begin{equation}
    \begin{split}
        \mathcal{L}_k = \sum_{j=1}^{d_h}{\left| \hat{H}^k_j - \hat{H}^t_j\right|}^2,
    \end{split}
\end{equation}
where $d_h$ is the hidden size of BERT.
\subsection{Training and Inference}
In this setting, we use a softmax classifier to predict label ${r}$ from the predefined set of classes $\mathcal{R}$ for a dialogue $\mathcal{D}$ with a pair of argument $(a_1,a_2)$. We concatenate the hidden states $h^{CLS1}$, $h^{CLS2}$ and the fused trigger feature $\hat{H}^t$ as the input, and we use the cross-entropy loss to train the classifier:
\begin{equation}  
    \begin{split}
        x&=(h^{CLS1}\oplus \hat{H}^{t} \oplus h^{CLS2}),\\
        p(r_i|\mathcal{D}, (a_1,a_2)) &= softmax(W_xx+b_x),\\
        {r'} &= \mathop{\arg\max}_{r_i} \ \ p(r_i|\mathcal{D}, (a_1, a_2)),\\
        \mathcal{L}_{r}&= CELoss({r'},r),
    \end{split}
\end{equation}
where $W_x \in {\mathbb{R}^{n \times 3d_h}},\; b_x \in {\mathbb{R}^n}$ are the weight matrix and bias, $n$ is the number of relation types, and $\oplus$ denotes the concatenation operator, ${r'}$ and $r$ are the predicted relation and relation label.

Finally, the relation classification task, the trigger span extraction task, and the knowledge-guidance task are trained in an end-to-end manner:
\begin{equation}
    \mathcal{L}_{total} = \lambda_r \mathcal{L}_{r} + \lambda_t \mathcal{L}_{t} + \lambda_k \mathcal{L}_{k},
\end{equation}
where $\lambda_r$, $\lambda_t$ and $\lambda_k$ are hyperparameters of weight factor.
\section{Experiments}\label{Section_experiments}
In this section, we first introduce the experiment settings and then report the experimental results as well as the discussion part, i.e., ablation studies and case studies.

\subsection{Experiment Settings}
\subsubsection{Dataset and Evaluation Metric}
We conduct experiments on the DialogRE dataset~\cite{DRE}, the first human-annotated DRE dataset, originating from the complete transcripts of the series \emph{Friends}. DialogRE has 36 relation types, 1788 dialogues, and 8119 argument pairs. The statistics of DialogRE are shown in Table~\ref{dataset}.
\begin{table}[!ht]
\centering
\renewcommand\arraystretch{1.2}
\caption{DialogRE Dataset Statistics.}
\label{dataset}
\begin{tabular}{llll} 
\hline
DialogRE                & Train & Dev   & Test   \\ 
\hline
\# Conversations         & 1073  & 358   & 357    \\
Average dialogue length & 229.5 & 224.1 & 214.2  \\
\# Argument pairs        & 5963  & 1928  & 1858   \\
Average \# of turns     & 13.1  & 13.1  & 12.4   \\
Average \# of speakers  & 3.3   & 3.2   & 3.3    \\
\hline
\end{tabular}
\end{table}
We report the macro F1-score of the proposed TLAG and all the baselines in both the standard and dialogue settings~\cite{DRE}, denoted as $F1$ and $F1_c$ respectively in the following sections. $F1_c$ is calculated by taking in the part of dialogues as input, instead of considering the entire dialogue.

\subsubsection{Implementation Details}
We use the BERT base model to initialize the shared encoder. We insert special tokens for speakers and we truncate a sequence when the input sequence length exceeds 512. We leverage the AdamW optimizer~\cite{adamw} with a learning rate of 3e-5. The weights of three loss functions are set as $\lambda_r = 1.0,\; \lambda_t = 0.3,\; \lambda_k = 0.1\;$. The batch size is 12. Experiments are conducted on a single Geforce GTX 3090 GPU. Our model is implemented in Pytorch. 

\subsubsection{Baseline Models}
For a comprehensive performance evaluation, we compare the proposed TLAG model with four types of existing models on the DialogRE dataset.
\begin{itemize}
    \item \textit{Sentence-level RE models} are applied to solve sentence-level relation extraction, mainly including neural network approaches like CNN~\cite{CNN}, LSTM~\cite{LSTM}, and BiLSTM~\cite{BiLSTM}. 
    
    \item \textit{Sequenced-based models} usually utilize pre-trained language models to encode utterances in the dialogue, including BERT~\cite{BERT}, BERTs~\cite{DRE}, SpanBERT~\cite{SpanBERT} and RoBERTa~\cite{RoBERTa}. BERTs is speaker-aware BERT. SpanBERT is an extended BERT that is trained by span prediction. RoBERTa is a retraining version of BERT with an improved training methodology.
    
    \item \textit{Graph-based models} include DHGAT~\cite{DHGAT}, LSR~\cite{LSR}, GDPNet~\cite{GDPNet} and TUCORE~\cite{TUCORE}. DHGAT constructs a heterogeneous graph over a dialogue. LSR induces the latent dependency structure. GDPNet develops a multi-view latent graph to better capture the key features. TUCORE constructs a turn-level graph to capture the relational information.
    
    \item \textit{Trigger-enhanced models} include TWP~\cite{Trigger_Words_Prediction} and TREND~\cite{TREND}. These models leverage trigger information from the training set. TWP proposes trigger word prediction as an auxiliary task. TREND proposes a trigger span prediction task and leverages the trigger feature for relation classification.
\end{itemize}

\begin{table}[!htbp]
\centering
\caption{Main results on DialogRE dataset. $\dagger$ means the results are reproduced by us.}
\label{main_results}
\renewcommand\arraystretch{1.2}
\begin{tabular}{llllll} 
\hline
\multirow{2}{*}{Type}             & \multirow{2}{*}{Model} &  \multicolumn{2}{c}{DialogRE-Dev}   & \multicolumn{2}{c}{DialogRE-Test}  \\ 
\cline{3-6}
                                  &                        & $F1$            & $F1_c$         & $F1$            & $F1_c$          \\ 
\hline
\multirow{3}{*}{Sentence-level}   & CNN                    & 46.1          & 43.7          & 48.0          & 45.0           \\
                                  & LSTM                   & 46.7          & 44.2          & 47.4          & 44.9           \\
                                  & BiLSTM                 & 48.1          & 44.3          & 48.6          & 45.0           \\ 
\hline
\multirow{4}{*}{Sequence-based}   & BERT                   & 60.6          & 55.4          & 58.5          & 53.2           \\
                                  & BERTs                  & 63.0          & 57.3          & 61.2          & 55.4           \\
                                  & SpanBERT               & 64.6          & 58.8          & 61.8          & 55.8           \\
                                  & RoBERTa                & 65.2          & 61.4          & 62.8          & 58.8           \\ 
\hline
\multirow{4}{*}{Graph-based}      & DHGAT                  & 60.2          & 57.1          & 59.9          & 55.8           \\
                                  & LSR                    & 62.8          & 58.4          & 61.4          & 56.2           \\
                                  & GDPNet                 & 67.1          & 61.5          & 64.9          & 60.1           \\
                                  & TUCORE             & 66.8          & 61.0          & 65.5          & 60.2           \\ 
\hline
\multirow{4}{*}{Trigger-enhanced} & TWP                    & 66.8          & 61.5          & 65.5          & 60.5           \\
& TREND & - & - &66.8&-  \\
    & TREND$^\dagger$                  &67.4          & 61.8        & 65.8          & 60.4           \\
                                  & Ours                   & \textbf{68.1} & \textbf{61.9} & \textbf{66.6} & \textbf{60.8}  \\
\hline
\end{tabular}
\end{table}

\subsection{Main Results}
 As shown in Table~\ref{main_results}, our TLAG model achieves 66.6\% $F1$ and 60.8\% $F1_c$ on the test set, which surpasses the baseline model in both validation and test sets, demonstrating the effectiveness of our approach. The detailed comparisons and discussions are given as follows.

\begin{itemize}
    \item \textit{Compared with sentence-level RE models.} Our TLAG outperforms baseline models CNN, LSTM, and BiLSTM by more than 15\% in $F_1$ and $F1_c$ on the test set. DRE requires reading multiple sentences in a dialogue to infer the relations of the argument pairs by considering all the dialogue information, making it difficult for sentence-level models to effectively reason relation across utterances.
    
     \item \textit{Compared with sequenced-based models.}  Our TLAG outperforms these sequence-based models. These models treat the dialogue as a long sequence and ignore the structural information within the dialogue, which results in low performance.
    
    \item \textit{Compared with graph-based models.} Our TLAG is still able to get the best performance, giving significant improvement on the DialogRE dataset. Although the graph-based models take into account the particular characteristics of the dialogue text, the low information density issue is still not well resolved.
    
    \item \textit{Compared with trigger-enhanced models.} Our TLAG consistently performs the best among all the trigger-enhanced models. We observe that our model achieves 0.4\% and 0.3\% improvement over TREND and TWP respectively in terms of $F1_c$ on the DialogRE test set. The comparisons confirm the effectiveness of our model in capturing the triggers for DRE. Although the recent strong baselines TWP~\cite{Trigger_Words_Prediction} and TREND~\cite{TREND} also induce a trigger span predictor to capture trigger words in the dialogue, without the adaptive trigger fusion mechanism and the label-aware knowledge leads to much lower performance than our method.
\end{itemize}

\begin{table}[!ht]
\centering
\caption{Ablation Results on DialogRE.}
\label{ablation}
\renewcommand\arraystretch{1.2}
\begin{tabular}{lcc} 
\hline
\multirow{2}{*}{Model} & \multicolumn{2}{c}{DialogRE-Test}  \\ 
\cline{2-3}
                       & $F1$   & $F1_c$                  \\ 
\hline
Ours                   & \textbf{66.6} & \textbf{60.8}                   \\

 \quad w/o Adaptive Trigger Fusion        & 66.3 & 60.7                   \\

 \quad w/o Label-aware Knowledge guidance           & 66.2 & 60.6                   \\
TREND                  & 65.8 & 60.4                   \\
TWP                    & 65.5 & 60.5                   \\
\hline
\end{tabular}
\end{table}

\subsection{Ablation Studies}
We also perform ablation experiments to better understand the importance of each part of our proposed method. The ablation results are reported in Table \ref{ablation}. First, we replace the adaptive trigger fusion module with mean pooling, we have observed that $F1$ and $F1_c$ scores decrease to 66.3\% and 60.7\% on the test set. This proves that our adaptive trigger fusion effectively synthesizes trigger features to the model. Next, we remove the label-aware knowledge during training, the performance has a 0.4\% drop in $F1$ score and a 0.2\% drop in $F1_c$ score. This result indicates that label-aware knowledge guidance plays an important role. Notably, compared with baseline models which also utilize the trigger, our model still performs better after removing one module.

\begin{table}[!ht]
\begin{center}
\renewcommand\arraystretch{1.2}
\caption{Case Studies.}
\label{case_study}
\begin{tabular}{l|l}
\hline \textbf{Case 1} & \textbf{S1:} ... My \textcolor{red}{friend} Susan is so funny... \\ &\textbf{S3:} Hi! Hi! You’re crazy! Okay? This is Emily.  \\
 & ... \\
& \textbf{S1:} ... we thought Carol was straight before I \textcolor{blue}{married} her!
 \\ &\textbf{S4:} Yeah, I definitely. I don’t like the name Ross.
 \\ \hline
\textbf{BERTs} & (\textbf{S1}, \texttt{per:friends}, \textbf{Carol}) \xmark   \\
\textbf{TREND} & (\textbf{S1}, \texttt{per:friends}, \textbf{Carol}) \xmark  \,\,\,\,\,\,\,\,\,\,\, trigger: \textcolor{red}{friend}  \,\, \xmark \\
\textbf{TLAG} & (\textbf{S1}, \texttt{per:spouse}, \textbf{Carol}) \cmark  \,\,\,\,\,\,\,\,\,\, trigger: \textcolor{blue}{married} \cmark  \\ \hline
\hline \textbf{Case 2} & \textbf{S1:} This is unbelievable Phoebs, how can you be married? \\ &\textbf{S2:} I'm not married, ya know, he's just a friend… \\ &\textbf{S3:} I can't believe you \textcolor{blue}{married} Duncan. I mean how could  \\ & \quad \ \ you not tell me? We \textcolor{red}{lived together}, we told each other \\& \quad \ \ everything.\\
\hline
\textbf{BERTs} & (\textbf{S1}, \texttt{per:spouse}, \textbf{Duncan}) \cmark   \\
\textbf{TREND} & (\textbf{S1}, \texttt{per:roommate}, \textbf{Duncan}) \xmark \  trigger: \textcolor{red}{lived together} \xmark \\
\textbf{TLAG} & (\textbf{S1}, \texttt{per:spouse}, \textbf{Duncan}) \cmark \quad \, trigger: \textcolor{blue}{married} \cmark \\  \hline
\end{tabular}
\end{center}

\label{tab:badcase}
\end{table}

\subsection{Case Studies}
The dataset statistics show that $49.6\%$ of argument pairs have triggers~\cite{DRE} in DialogRE. Therefore, it is crucial that the model can predict the triggers correctly. We give some case studies to analyze the results produced by our approach and the baseline models. Results in Table~\ref{case_study} show that our approach is able to predict the trigger words and relation correctly. In case 1, BERTs and TREND both identify the relation as a wrong type “\textit{per: friends}”. BERTs ignores the trigger information. TREND fails to predict the correct trigger. Our method correctly recognizes the trigger word “\textit{married}” between S1 and S2, which promotes the right result. In case 2, BERTs predicts the relation “\textit{per: spouse}” correctly. However, TREND obtains the wrong trigger “\textit{lived together}", which brings noise to the model and predicts a wrong relation “\textit{per: roommate}” between S1 and “\textit{Duncan}”. Contrarily, our method is able to leverage the trigger information efficiently and infer the correct relation in a such complex scenario.


\section{Conclusion}\label{section_conclusion}

In this work, we propose TLAG, a novel dialogue-based relation extraction model. Compared with the baseline methods, TLAG improves performance by fully mining trigger information and using label-aware knowledge guidance. We further conduct a set of ablation studies to verify the importance of the proposed adaptive trigger fusion mechanism and label-aware knowledge guidance module. Besides, the results of case studies also show the effectiveness of our TLAG. In the future, we will focus on how to leverage domain knowledge graph to improve cross-domain dialogue-based relation extraction.

\section*{Acknowledgment}
This paper was partially supported by Shenzhen Science \& Technology Research Program (No: GXWD2020123116580
7007-20200814115301001) and NSFC (No: 62176008).


\bibliographystyle{IEEEtran}
\bibliography{reference}

\end{document}